\documentclass[runningheads]{llncs}
\usepackage[T1]{fontenc}
\usepackage{graphicx}
\usepackage{caption}
\usepackage{subcaption}
\usepackage{csquotes}
\usepackage{colortbl}

\usepackage[hidelinks]{hyperref}

\usepackage{xcolor}

\definecolor{darkred}{RGB}{186,0,0}
\definecolor{darkblue}{RGB}{0, 0, 186}

\newcommand{\higherval}{\cellcolor{red!12}\textcolor{darkred}{$>$}}
\newcommand{\lowerval}{\cellcolor{blue!12}\textcolor{darkblue}{$<$}}

\begin{document}
\title{A Study of Augmentation Methods for Handwritten Stenography Recognition}

% AUTHORS AND INSTITUTIONS COMMENTED OUT FOR ANONYMITY
\author{Raphaela Heil\orcidID{0000-0002-5010-9149} \and
Eva Breznik\orcidID{0000-0003-3147-5626} }
%Third Author\inst{3}\orcidID{2222--3333-4444-5555}}
%%
\authorrunning{R. Heil, E. Breznik}
%% First names are abbreviated in the running head.
%% If there are more than two authors, 'et al.' is used.
%%
\institute{Centre for Image Analysis,\\ Department of Information Technology, \\Uppsala University, Uppsala, Sweden\\
\email{\{firstname\}.\{lastname\}@it.uu.se}\\
}
\maketitle              % typeset the header of the contribution

\newcommand{\todo}[1]{\textbf{TODO: #1}}
\begin{abstract}
One of the factors limiting the performance of handwritten text recognition (HTR) for stenography is the small amount of annotated training data. To alleviate the problem of data scarcity, modern HTR methods often employ data augmentation. However, due to specifics of the stenographic script, such settings may not be directly applicable for stenography recognition. In this work, we study 22 classical augmentation techniques, most of which are commonly used for HTR of other scripts, such as Latin handwriting. Through extensive experiments, we identify a group of augmentations, including for example contained ranges of random rotation, shifts and scaling, that are beneficial to the use case of stenography recognition. Furthermore, a number of augmentation approaches, leading to a decrease in recognition performance, are identified. Our results are supported by statistical hypothesis testing. Links to the publicly available dataset and codebase are provided.

%The abstract should briefly summarize the contents of the paper in 150--250 words.

\keywords{Stenography  \and Handwritten text recognition \and CNNs \and Augmentation study}
\end{abstract}

\section{Introduction}

Deep learning-based approaches form the majority of the state-of-the-art methods for handwritten text recognition (HTR) at the time of writing. While these approaches have been shown to reach high levels of recognition performance, they typically require large amounts of annotated data during training. This can for example pose a challenge to HTR for historic manuscripts. Here the acquisition of annotated data can be costly and time-consuming, as it often requires trained professionals, such as historians or palaeographers. Data availability may further be limited by the use of a rarely-used script or language, as is for example the case for the Khmer language \cite{khmer}.

When the acquisition of additional training data is not feasible, or possible, a commonly used approach for artificially increasing the dataset size is to use data augmentation \cite{hinton,elastic}. Here, slight alterations, for example rotation and scaling, are applied to the images in order to increase the visual variety. 

\begin{figure}[t]
    \centering
    \includegraphics[width=0.7\textwidth]{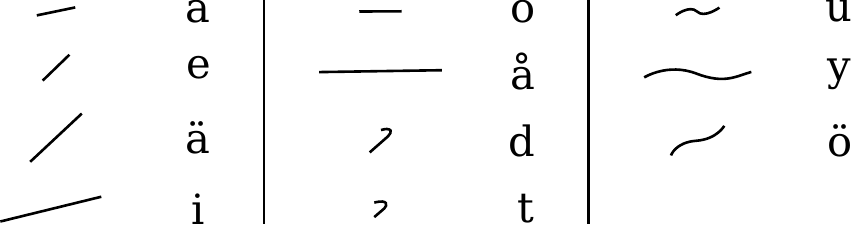}
    \caption{Selected examples of symbols and respective character transliterations from the Swedish \enquote{Melin} stenography system.}
    \label{fig:steno}
\end{figure}

%With 
In the particular case of stenography, the acquisition of more data is limited both by the special skill required to transliterate the writing, as well as the limited use of the script, making data augmentation options especially interesting.  %a  
In contrast to scripts like Latin, stenography (also called shorthand) typically uses short strokes to represent characters, n-grams or even whole words. As can be seen in the excerpt of the Swedish stenography alphabet in \autoref{fig:steno}, features like rotation and scale play a considerable role in differentiating between certain symbols. This raises the question whether typical HTR augmentation techniques, which often include changes in rotation and scale, can also be applied to stenography or whether they may cause certain symbols to be interpreted as others (e.g. \enquote{a} as \enquote{e} and vice-versa).

In this work, we aim to address this question by studying the applicability of a selection of commonly used HTR augmentation techniques for the case of handwritten stenography recognition. We experiment with a deep learning architecture that has previously been shown to perform best for Swedish stenography \cite{lion}, and observe how different augmentation techniques affect the text recognition performance on the public LION dataset \cite{lion}, consisting of stenographic manuscripts, written by the Swedish children's book author Astrid Lindgren. 

\section{Related Work}
\subsection{Handwritten Stenography Recognition}
At the time of writing, very little research has been conducted on deep learning-based handwritten stenography recognition (HSR). In \cite{steno-classification1} and \cite{steno-classification2}, convolutional neural networks (CNNs) are used to classify a selection of words written in Pitman's, respectively Gregg's, stenography system. Both types of stenography are primarily used in English-language contexts.

Zhai et al.\ \cite{gregg1916} propose to use a CNN-based feature extractor, followed by a recurrent neural network to generate transliterations of individual words, written in Gregg's stenography. 

Lastly, our previous work \cite{lion} introduces the novel LION dataset, based on the Swedish stenography system Melin, using excerpts from Astrid Lindgren's manuscripts. It furthermore establishes a baseline performance using the model proposed by Neto et al.\ \cite{flor}. To the best of our knowledge, LION is currently the only available line-based dataset for any system of stenography. 

\subsection{Data Augmentation Methods for Handwritten Text Recognition}
A variety of augmentation approaches for HTR has been proposed in the literature. Some of the most commonly used augmentations include rotations, translations, shearing and scaling, as well as greyscale dilations and erosions %, which are for example employed in 
\cite{pylaia,flor,tomas}. Similarly to these, Retsinas et al.\ \cite{htr-best-practices} propose the use of rotations, skewing and additionally applying Gaussian noise. 

Wick et al.\ \cite{rostock} also use greyscale dilation and erosion and further simulate the variability in handwriting by applying grid-like distortions, originally proposed by Wigington et al.\ \cite{distortion-aug}. Krishnan and Jawahar \cite{hwnetv2} follow a similar approach, applying translation, scaling, rotation and shearing and combine these with the elastic transformations, proposed by Simard et al.\ \cite{elastic}.

All of the aforementioned works apply each augmentation with an independent probability (often 0.5) to each image in the training set. Furthermore, the augmentation parameters, for example the angle of rotation, are sampled per image from a user-defined range, %(e.g. $[-3, 3]$ degrees), 
yielding a wide variety of altered datasets.

Wilkinson et al.\ \cite{ctrlf} use the same augmentation approach and parameters as \cite{tomas} but arrange the perturbed images into lines and whole pages, thereby creating artificial data for segmentation-free word spotting. 

Zhai et al.\ \cite{gregg1916} apply similar transforms to all of the above but do not sample the respective parameters. Instead, they use fixed, pre-defined values and generate a total of eight images (with fixed choice of augmentation parameters) for every input.

%another form of augmentation: synthetic data
Instead of perturbing the available data, a different line of augmentation techniques is centred around creating entirely synthetic datasets.
Circumventing the use of the data at hand entirely, Krishnan et al.\ \cite{hw-synth} generate words by using publicly available handwriting fonts. This approach is currently not applicable in the case of stenography, as such a library of handwritten symbols and fonts is not available for these scripts, to the best of our knowledge. 

While Alonso et al.\ \cite{gan-messina}, Kang et al.\ \cite{gan-writing} and Mattick at al.\ \cite{mattick} also propose to generate synthetic data, they employ generative adversarial networks to synthesise words in the style of a given dataset or sample image. To the best of our knowledge, approaches in this line of research currently employ word-level annotations (segmentation and transliteration), which are not available for the dataset in our study.

\section{Study Design}
In order to investigate the effect of different augmentations on the text recognition performance, we evaluate a variety of image transformations that are applied during training. We compare the results with those obtained from a baseline model, trained on the original data, without any augmentations.

\begin{table}[h]
    \centering
    \caption{Summary of augmentation configurations and the names by which they are referred to in this paper.}
    \begin{tabular}{c|c|c}
        Name & Augmentation Type & Parameters \\
        \hline
        baseline & none & N/A\\
        \hline
        rot1.5 & Random Rotation & $[-1.5, 1.5]$ degrees\\
        rot5 & Random Rotation & $[5, 5]$ degrees\\
        rot10 & Random Rotation & $[10, 10]$ degrees\\
        positive & Random Rotation & $[0, 1.5]$ degrees\\
        negative & Random Rotation & $[-1.5, 0]$ degrees\\
        rot+2 & Fixed Rotation & 2 degrees\\
        rot-2 & Fixed Rotation & -2 degrees\\
        \hline
        square-dilation & Random Dilation & square SE, $[1..4]$ px \\
        disk-dilation & Random Dilation & disk SE, $[1..4]$ px\\
        square-erosion & Random Erosion & square SE, $[1..3]$ px\\
        disk-erosion & Random Erosion & disk SE, $[1..3]$ px\\
        \hline
        shift & Random Shift & $horiz.=[0, 15]$, $vert.=[-3.5, 3.5]$\\
        elastic & Random Elastic Transform. \cite{elastic} & $\alpha=[16, 20]$, $\sigma=[5, 7]$\\
        shear & Random Horizontal Shearing & $[-5, 30]$ degrees\\
        shear30 &Random Horizontal Shearing & $[-30, 30]$ degrees\\
        scale75 & Random Scaling & $[0.75, 1]$\\
        scale95 & Random Scaling & $[0.95, 1]$\\
        \hline
        mask10 & Random Column Masking & 10\% of columns\\
        mask40 & Random Column Masking & 40\% of columns\\
        noise & Gaussian Noise & $\sigma=\{0.08, 0.12, 0.18\}$\\
        dropout & Pixel Dropout & $[0, 20]$\% of pixels\\
        blur & Gaussian Blur & kernel=5, $\sigma=[0.1, 2]$\\
    \end{tabular}
    
    \label{tab:param_summary}
\end{table}

\subsection{Examined Augmentations}
Below, the examined augmentations and the respective parameter configurations are briefly presented. Each experiment configuration is denoted with a name by which it will be referred to in the remainder of this paper. A summary of names, augmentation types and parameters is shown in \autoref{tab:param_summary}. Additionally, the connected supplementary \cite{suppl} contains a visualisation of the impact of selected augmentation parameters on the original line image.

The implementations for most of the examined augmentations were provided by \cite{imgaug}, \cite{imagecorruptions} and \cite{torchvision}. All others are available in our project repository (cf. \autoref{sec:availability} - \nameref{sec:availability}).

\subsubsection{Baseline} 
In order to establish a baseline performance, the model is trained on the original data, not using any augmentations. 

\subsubsection{Rotations}
We examine a variety of rotations, largely based on configurations used in related text recognition works. Three models are trained with random rotations in the ranges $[-1.5, 1.5]$ (\enquote{rot1.5}) \cite{flor}, $[-5, 5]$ (\enquote{rot5}) \cite{tomas} and $[-10,10]$ (\enquote{rot10}). Following \cite{gregg1916}, one model each is trained with a fixed rotation of +2, respectively -2, degrees (\enquote{rot+2}, \enquote{rot-2}). Lastly, one model each is trained with only positive, respectively negative rotations, in the ranges $[0, 1.5]$ (\enquote{positive}) and $[-1.5, 0]$ (\enquote{negative}).

\subsubsection{Morphological Operations}\label{sec:aug_morph} 
With respect to morphological operations, we consider greyscale dilations and erosions with a square structuring element (SE) \cite{pylaia,flor,tomas}. Additionally, we evaluate the use of a disk SE, because the primary writing implement in the LION dataset is a pencil, which typically features a round footprint. We refer to these augmentations as \enquote{square-dilation}, \enquote{disk-dilation}, \enquote{square-erosion} and \enquote{disk-erosion}, respectively. For each application of these augmentations (i.e. per image) the size of the SE is sampled from $[1..3]$ for erosions and $[1..4]$ for dilations. It should be noted that in the case of a square SE, the size refers to the width, whereas it indicates the radius for disks.

\subsubsection{Geometric Augmentations}In addition to the aforementioned rotations, we consider a number of other geometric augmentations. %\todo{I don't even know what to fucking call this properly ... } 

Concretely, we evaluate shearing with angles in the range of $[-5, 30]$ (\enquote{shear}, following \cite{tomas}) and $[-30, 30]$ (\enquote{shear30}). Furthermore, we examine downscaling with factors in the range of $[0.95, 1]$ (\enquote{scale95}, \cite{flor}) and $[0.75, 1]$ (\enquote{scale75}). The rescaled image is zero-padded at the top and bottom to the original size for further processing. 

Following Krishnan and Jawahar \cite{hwnetv2} we investigate the applicability of the elastic transformation (\enquote{elastic}), proposed by \cite{elastic}. In this work, we sample $\alpha$ from the range $[16, 20]$ and $\sigma$ from $[5, 7]$. 

Lastly, we consider random shifts (\enquote{shift}) in horizontal and vertical direction, within the pixel ranges of $[0, 15]$, respectively $[-3.5, 3.5]$. For both directions, non-integer shifts may occur.

\subsubsection{Intensity Augmentations} Finally, we consider a selection of augmentations that affect an image's pixel intensities. We examine both pixel dropout (\enquote{dropout}), i.e. zero-masking of random pixels, with rates in the range of $[0, 20]\%$ of the image, and random column masking with fixed rates of $10\%$ (\enquote{mask10}) and $40\%$ (\enquote{mask40}) of columns. Additionally, we consider the application of Gaussian noise (\enquote{noise}), with $\sigma=\{0.08, 0.12, 0.18\}$, and Gaussian blurring (\enquote{blur}) with a kernel size of 5 and $\sigma=[0.1, 2]$.

\subsection{Dataset}
This study is centred around the LION dataset \cite{lion}, which is based on stenographed manuscripts by Swedish children's book author Astrid Lindgren (1907 – 2002). It consists of 198 pages, written in Swedish, using the \textit{Melin} shorthand system. For each page, line-level bounding boxes and transliterations are available, the latter of which were provided by stenographers through expert crowdsourcing \cite{dhnb2022}. Overall, the dataset has an alphabet size of 51, entailing the 26 lower-case Latin characters (a-z), the Swedish vowels \enquote{äöå}, digits 0-9, and a selection of punctuation marks and quotes appearing in the text. 

\begin{figure}[b]
\includegraphics[width=\textwidth]{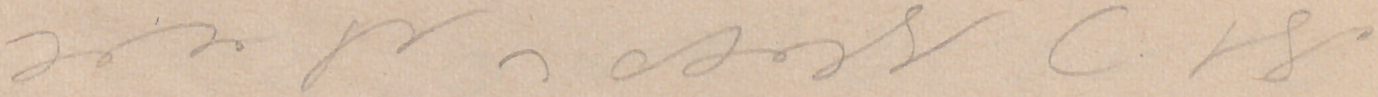}
\includegraphics[width=\textwidth]{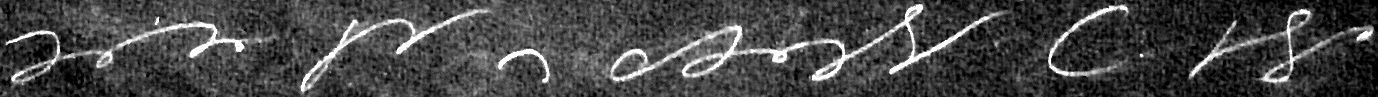}
\caption{Example image from the LION dataset \cite{lion}. Top: original line image, bottom: preprocessed line image. The transliteration reads: \enquote{jonatan hette inte lejonhjärta från början} (English: \textit{jonatan was not called lionheart from the beginning}).} \label{fig:line-examples}
\end{figure}

We follow the datasplitting proposed by \cite{lion}, resulting in a five-fold cross-validation set of 1530 lines (306 lines per fold). This portion of the dataset contains texts from \textit{The Brothers Lionheart} (chapters 1 - 3, 5, 6; original title: Bröderna Lejonhjärta). The test split consists of 474 in-domain lines, covering chapter four of \textit{The Brothers Lionheart}, and 191 out-of-domain lines, containing portions from other literary works by Lindgren. All lines in the dataset are preprocessed, following the method proposed by \cite{lion}.  For concrete implementation details, the interested reader is referred to our implementation (cf- \autoref{sec:availability}). \autoref{fig:line-examples} shows an example line from the dataset in its original form (top) and the preprocessed version (bottom).

\subsection{Model Architecture}
We focus our study on the deep neural network architecture proposed by Neto et al., who demonstrate that this model performs well in limited-resource line-based HTR settings \cite{flor}. This architecture outperformed other connectionist temporal classification (CTC) \cite{ctc_original,ctc_liwicki} architectures in preliminary experiments \cite{lion}. The model consists of a convolutional block, mixing regular and gated convolutions \cite{gated-cnn}, for feature extraction, followed by a recurrent block, employing bi-directional gated recurrent units \cite{gru}, which generate the output sequence.  

Further details regarding the model architecture can be found in our PyTorch \cite{pytorch} implementation in the accompanying code (cf. \autoref{sec:availability} - \nameref{sec:availability}).

\subsection{Experimental Settings}

All of the experiments below follow the same general procedure and only differ in regard to the augmentation that is applied to the data during training. In each epoch, each image is augmented at a rate of 50\%. Afterwards, all images are scaled to a fixed height of 64 pixels and padded with the background colour (black) to a width of 1362 pixels. Target sequences are padded with the token indicating blanks in the CTC-loss (here zero), to a length of 271. 

We follow the same training protocol as \cite{lion}, which is included for reference in the supplementary material \cite{suppl}.

Each augmentation configuration is trained 30 separate times for each of the five cross-validation folds, yielding a total of 150 model checkpoints (i.e. sets of weights). All checkpoints are evaluated by measuring the character error rate (CER) and word error rate (WER) for the transliterations, obtained via best path decoding \cite{ctc_original,ctc_liwicki} on the test set. We do not consider advanced decoding techniques, such as word beam search \cite{wbs} or language models, in order to rule out any effects that these may have on the obtained transliterations. CER and WER are defined as:

\begin{equation}
ErrorRate = \frac{S + D + I}{N}
\end{equation}

where $S$ is the number of character (word) substitutions, $D$ the number of character (word) deletions and $I$ the number of character (word) insertions, that are required to convert a given sequence to a reference sequence. $N$ denotes the number of characters (words) in the reference string. 

Final statistical testing for significant differences is performed by a Wilcoxon signed-rank test \cite{wilcoxon} on the mean results, with a Bonferroni correction \cite{bonferroni} for multiplicity.

\section{Results and Discussion}

\begin{figure}[b]
    \centering
    \begin{subfigure}[b]{0.30\textwidth}
         \centering
    \includegraphics[height=20px]{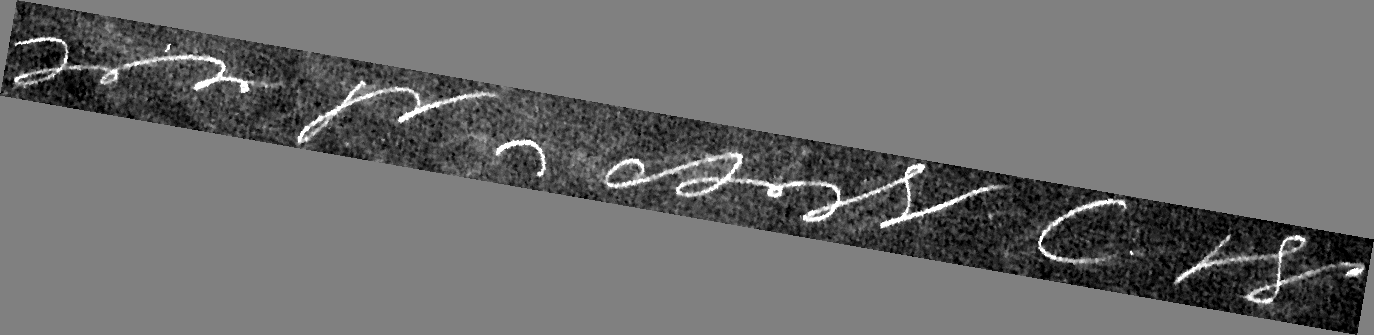}
         \caption{-10 degrees}
     \end{subfigure}
     \hfill
     \begin{subfigure}[b]{0.65\textwidth}
         \centering
    \includegraphics[height=20px]{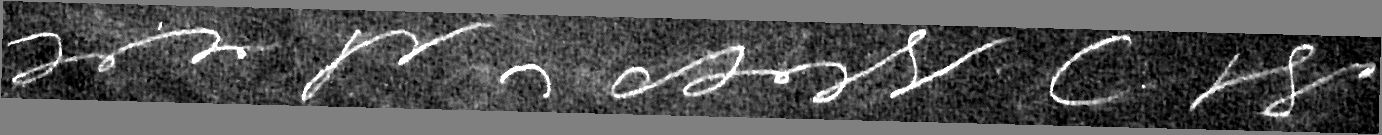}
         \caption{-1.5 degrees}
     \end{subfigure}
    \caption{Examples for the impact of different rotations on the original image content size. Padding value set to grey to emphasise the required padding.}\label{fig:rot_example}
\end{figure}

The mean error rates for the 22 different augmentations, as well as for the augmentation-free baseline, are shown in \autoref{tab:summary}. As can be seen, the augmentations \textit{rot10}, \textit{square-erosion}, \textit{disk-erosion}, \textit{square-dilation}, \textit{disk-dilation}, \textit{mask40} and \textit{blur} all result in significant decreases in performance. 

For \textit{rot10}, this result is likely attributable to the extreme rescaling of the line contents that occurs when height-normalising the rotated image. In order to avoid cutting off parts of a rotated line, the image size is expanded accordingly and padded with black. Larger rotations require a larger target image to accommodate the whole line, resulting in a smaller text size, as compared to smaller rotations, when scaled to the same height. \autoref{fig:rot_example} shows an example of the impact on the image content, when rotating the original line image to by 10, respectively 1.5 degrees.

Regarding the increase in error rates for both kinds of erosions, it can be observed that these augmentations thin the original strokes considerably. We argue that the selected erosions thin the strokes too much, leaving too little information for the recognition behind. This argument is supported by the significantly lower ($p<0.01$) performance of the disk-erosion, as compared to the square one. As indicated earlier (cf. \autoref{sec:aug_morph}), the disk SEs are larger than their square counterparts, thus thinning the strokes more strongly and leaving even less information behind. It should be noted here, that this is not an issue that can be directly attributed to stenography but rather to the use of a relatively thin pencil, which results in thin strokes that are more susceptible to erosions.

\begin{figure}[h]
    \centering
    \begin{subfigure}[b]{0.4\textwidth}
         \centering
    \includegraphics[height=20px]{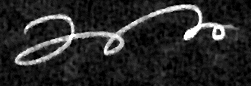}
         \caption{original}
     \end{subfigure}
     \hfill
     \begin{subfigure}[b]{0.4\textwidth}
         \centering
    \includegraphics[height=20px]{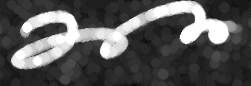}
         \caption{disk-dilation, radius $= 4$px}
     \end{subfigure}
    \caption{Example demonstrating the filling of several loops due to a dilation with a large structuring element.}\label{fig:dilation}
\end{figure}

Considering the two dilation-based augmentations, it can firstly be noticed that, in contrast to the WER, no significant difference for the CER can be determined in the case of the \textit{square-dilation}. This may be attributable to large dilations closing the gap between words. Such word boundary errors will only marginally affect the CER as they require a single character substitution or insertion to establish the gap. They will however lead to considerable increases in the WER, as one word substitution and one word insertion will be required to recover the recognition mistake. This may also be a contributing factor for the \textit{disk-dilation}. In this regard it can however also be observed that the disk SE, whose footprint is larger than the square's, leads to the loss of details, regarding small, and long and thin, loops, which are frequently used in Melin's stenography system. This loss of definition, an example of which is shown in \autoref{fig:dilation}, may lead to incorrect transliterations.

Since the performance of \textit{mask10} is on par with the baseline, a probable  explanation for the decrease in performance when using \textit{mask40} is that the level of masking is simply too high, again removing too much of the information and thus affecting the recognition performance. The decrease in performance for the \textit{blur} augmentation may be attributable to the loss of definition between smaller symbols, like \enquote{a} and \enquote{o}, and \enquote{u} and \enquote{ö}.  

For six of the augmentations, namely rot5, rot+2, rot-2, mask10, noise and dropout, no differences can be observed with respect to the baseline. While including these may not directly harm the performance, they also do not appear to contribute to the learning and are therefore not of immediate interest to the concrete recognition task.

Finally, ten of the examined augmentations yield significant decreases in error rates, as compared to the baseline. Considering the improvements obtained when applying conservative rotations, in the overall range of $[-1.5,1.5]$ (\textit{rot1.5}, \textit{positive}, \textit{negative}), we cannot find any indications supporting the initial concern that rotations in general may negatively affect the transliteration of certain characters. Obviously, the angles between the similar symbols, for example \enquote{a} and \enquote{e}, tend to be much larger than the ones examined in these experiments. However, as discussed for the case of rot10, even rotations closer to the actual angle between similar characters are likely to lead to scaling related issues, outweighing any potential recognition errors caused directly by rotated symbols. 

Regarding the decrease in error rates for the remaining six augmentations, \textit{shift}, \textit{shear}, \textit{shear30}, \textit{elastic}, \textit{scale75} and \textit{scale95} it can be noted that all of these correspond to variations that, to some degree, naturally occur in the dataset. Applying these augmentations therefore constitutes plausible transformations within the dataset domain and increases the visual variety of the training data. % is this sentence too wild? No. Makes it clear that is not necessarily stenography related specifically

Given the reduction in error rates when applying several of the augmentations individually, the follow-up question arose whether a combination of some of these would have a similar positive effect. We therefore selected the three best performing augmentations, \textit{rot1.5}, \textit{shift} and \textit{scale75}, and repeated the experiment with a combination of these. Here, each augmentation in the sequence is applied with an independent probability of $0.5$. This configuration achieves significantly lower error rates than the baseline, with a  mean CER of \textbf{0.3090} (0.0208) and WER of \textbf{0.5603} (0.0253). 

\begin{table}[h]
    \centering
    
   \caption{Mean (and std. dev.) CER and WER per augmentation type. For both metrics, lower values are better. Comparison columns indicate if the mean for the inspected augmentation type is significantly higher ($>$, red), lower ($<$, blue), or no different ($-$) than the baseline mean, according to the Wilcoxon signed-rank test ($p<0.01$, after applying a Bonferroni correction with $n=22$). }
    \label{tab:summary}
    \begin{tabular}{l|c|c|c|c}
Augmentation &    CER (Std.dev.) & CER Comparison&   WER (std.dev.)&WER Comparison\\
\hline
baseline & 0.3174 (0.0123) & N/A & 0.5715 (0.0145)& N/A\\
\hline
rot1.5 & 0.3048 (0.0298) & \lowerval& 0.5561 (0.0334)& \lowerval\\
rot5 & 0.3167 (0.0351) & $-$ & 0.5674 (0.0383) & $-$\\
rot+2 & 0.3114 (0.0203) & $-$& 0.5616 (0.0245) & \lowerval\\
rot-2 & 0.3134 (0.0314) & $-$ & 0.5640 (0.0367)& $-$\\
positive & 0.3066 (0.0236) & \lowerval& 0.5584 (0.0278) & \lowerval\\
negative & 0.3062 (0.0225)  & \lowerval & 0.5584 (0.0273)& \lowerval\\
rot10 & 0.3585 (0.0753) & \higherval & 0.6113 (0.0659)& \higherval\\
\hline
square-erosion & 0.3189 (0.0156) & \higherval& 0.5740 (0.0187)& \higherval\\
disk-erosion & 0.3352 (0.0568) & \higherval & 0.5899 (0.0390)& \higherval\\
square-dilation & 0.3182 (0.0135)&$-$  & 0.5739 (0.0173) & \higherval\\
disk-dilation & 0.3203 (0.0164)& \higherval & 0.5764 (0.0209)& \higherval\\
\hline
shift & 0.3065 (0.0126)& \lowerval  & 0.5583 (0.0162) & \lowerval\\
shear & 0.3103 (0.0131) & \lowerval& 0.5649 (0.0173)& \lowerval\\
shear30 & 0.3102 (0.0128) & \lowerval & 0.5641 (0.0157) & \lowerval\\
elastic & 0.3139 (0.0128)& \lowerval& 0.5675 (0.0165)& \lowerval\\
scale75 & 0.3040 (0.0190) & \lowerval& 0.5557 (0.0235)& \lowerval\\
scale95 & 0.3077 (0.0119)& \lowerval & 0.5614 (0.0156)& \lowerval\\
\hline
mask10 & 0.3182 (0.0147)& $-$& 0.5732 (0.0192) & $-$\\
mask40 & 0.3370 (0.0797)  & \higherval& 0.5875 (0.0540)& \higherval\\
noise & 0.3172 (0.0156) & $-$& 0.5715 (0.0187) & $-$\\
dropout & 0.3172 (0.0127) & $-$& 0.5727 (0.0159) & $-$\\
blur & 0.3195 (0.0153) & \higherval& 0.5758 (0.0196) & \higherval\\
\end{tabular}
\end{table}
% mixed & 0.3090 (0.0208) & 0.5603 (0.0253)\\

\section{Conclusions}
In this work we have examined 22 different augmentation configurations and their impact on handwritten stenography recognition. Based on the obtained results, we conclude that small rotations, shifting, shearing, elastic transformations and scaling are suitable augmentations for the recognition task and dataset at hand. Furthermore, an increase in performance could be observed for a combination of the top three augmentations, i.e. rotations in the range of $[-1.5,1.5]$ degrees, horizontal and vertical shifting by $[0,15]$ px, respectively $[-3.5,3.5]$ px, and scaling by a factor in the range of $[0.75,1]$. 

A decrease in performance can be observed for larger rotations ($\pm10$ degrees), which is attributable to the extreme rescaling resulting from the padding of augmented images. Besides this, the results obtained for erosions and dilations indicate that the examined configurations have adverse effects on the recognition performance. This can largely be attributed to the writing implement that was used in the dataset, highlighting that aspects like the stroke width should be taken into account when choosing augmentations for any kind of text recognition system.

Overall, initial concerns that commonly-used HTR augmentations may lead to confusions between certain symbols in the stenographic alphabet could not be confirmed.

Interesting avenues for future research include the creation of text lines by combining segmented words, for example as proposed by Wilkinson and Brun \cite{ctrlf}, as well as the investigation of text generation approaches similar to the ones proposed by Alonso et al.\ \cite{gan-messina}, Kang et al.\ \cite{gan-writing} and Mattick at al.\ \cite{mattick}.

\subsubsection{Data and Code Availability}\label{sec:availability}
The dataset and code will be available via Zenodo (\url{https://zenodo.org/}) with the camera-ready version of this paper.

\subsubsection{Acknowledgements}
This work is partially supported by Riksbankens Jubileumsfond (RJ) (Dnr P19-0103:1). The computations were enabled by resources provided by the National Academic Infrastructure for Supercomputing in Sweden (NAISS) at Chalmers Centre for Computational Science and Engineering (C3SE) partially funded by the Swedish Research Council through grant agreement no. 2022-06725. Author E.B. is partially funded by the Centre for Interdisciplinary Mathematics, Uppsala University, Sweden.

\bibliographystyle{splncs04}
\bibliography{main}

\begin{thebibliography}{10}
\providecommand{\url}[1]{\texttt{#1}}
\providecommand{\urlprefix}{URL }
\providecommand{\doi}[1]{https://doi.org/#1}

\bibitem{gan-messina}
Alonso, E., Moysset, B., Messina, R.: Adversarial generation of handwritten
  text images conditioned on sequences. In: 2019 International Conference on
  Document Analysis and Recognition (ICDAR). pp. 481--486 (2019)

\bibitem{dhnb2022}
Andersdotter, K., Nauwerck, M.: Secretaries at work: Accessing astrid
  lindgren's stenographed manuscripts through expert crowdsourcing. In:
  Berglund, K., Mela, M.L., Zwart, I. (eds.) Proceedings of the 6th Digital
  Humanities in the Nordic and Baltic Countries Conference {(DHNB} 2022),
  Uppsala, Sweden, March 15-18, 2022. vol.~3232, pp. 9--22 (2022)

\bibitem{bonferroni}
Bonferroni, C.: Teoria statistica delle classi e calcolo delle probabilita.
  Pubblicazioni del R Istituto Superiore di Scienze Economiche e Commericiali
  di Firenze  \textbf{8},  3--62 (1936)

\bibitem{gru}
Cho, K., van Merri{\"e}nboer, B., Bahdanau, D., Bengio, Y.: On the properties
  of neural machine translation: Encoder--decoder approaches. In: Proceedings
  of SSST-8, Eighth Workshop on Syntax, Semantics and Structure in Statistical
  Translation. pp. 103--111 (2014)

\bibitem{gated-cnn}
Dauphin, Y.N., Fan, A., Auli, M., Grangier, D.: Language modeling with gated
  convolutional networks. In: Precup, D., Teh, Y.W. (eds.) Proceedings of the
  34th International Conference on Machine Learning. Proceedings of Machine
  Learning Research, vol.~70, pp. 933--941. PMLR (06--11 Aug 2017)

\bibitem{ctc_original}
Graves, A., Fern\'{a}ndez, S., Gomez, F., Schmidhuber, J.: Connectionist
  temporal classification: Labelling unsegmented sequence data with recurrent
  neural networks. In: Proceedings of the 23rd International Conference on
  Machine Learning. p. 369–376. ICML '06, ACM, New York, NY, USA (2006)

\bibitem{ctc_liwicki}
Graves, A., Liwicki, M., Fernández, S., Bertolami, R., Bunke, H., Schmidhuber,
  J.: A novel connectionist system for unconstrained handwriting recognition.
  IEEE Transactions on Pattern Analysis and Machine Intelligence
  \textbf{31}(5),  855--868 (2009)

\bibitem{suppl}
Heil, R., Breznik, E.: {Supplementary Material} (2023),
  \url{https://uppsala.box.com/shared/static/d9o7s1cs282gzoki2ahz054ie0axq6lb.pdf},
  arXiv publ. in progress

\bibitem{lion}
Heil, R., Nauwerck, M.: Handwritten stenography recognition and the {LION}
  dataset. Manuscript  (2023)

\bibitem{imgaug}
Jung, A.B., Wada, K., Crall, J., Tanaka, S., Graving, J., Reinders, C., et~al.:
  {imgaug}. \url{https://github.com/aleju/imgaug} (2020), online; accessed
  01-Feb-2020

\bibitem{gan-writing}
Kang, L., Riba, P., Wang, Y., Rusi{\~{n}}ol, M., Forn{\'e}s, A., Villegas, M.:
  Ganwriting: Content-conditioned generation of styled handwritten word images.
  In: Vedaldi, A., Bischof, H., Brox, T., Frahm, J.M. (eds.) Computer Vision --
  ECCV 2020. pp. 273--289. Springer International Publishing, Cham (2020)

\bibitem{hw-synth}
Krishnan, P., Jawahar, C.V.: Matching handwritten document images. In: Leibe,
  B., Matas, J., Sebe, N., Welling, M. (eds.) Computer Vision -- ECCV 2016. pp.
  766--782. Springer International Publishing, Cham (2016)

\bibitem{hwnetv2}
Krishnan, P., Jawahar, C.V.: Hwnet v2: an efficient word image representation
  for handwritten documents. International Journal on Document Analysis and
  Recognition (IJDAR)  \textbf{22}(4),  387--405 (Dec 2019)

\bibitem{hinton}
Krizhevsky, A., Sutskever, I., Hinton, G.E.: Imagenet classification with deep
  convolutional neural networks. In: Pereira, F., Burges, C., Bottou, L.,
  Weinberger, K. (eds.) Advances in Neural Information Processing Systems.
  vol.~25. Curran Associates, Inc. (2012)

\bibitem{mattick}
Mattick, A., Mayr, M., Seuret, M., Maier, A., Christlein, V.: Smartpatch:
  Improving handwritten word imitation with patch discriminators. In:
  Llad{\'o}s, J., Lopresti, D., Uchida, S. (eds.) Document Analysis and
  Recognition -- ICDAR 2021. pp. 268--283. Springer International Publishing,
  Cham (2021)

\bibitem{imagecorruptions}
Michaelis, C., Mitzkus, B., Geirhos, R., Rusak, E., Bringmann, O., Ecker, A.S.,
  Bethge, M., Brendel, W.: Benchmarking robustness in object detection:
  Autonomous driving when winter is coming. arXiv preprint arXiv:1907.07484
  (2019)

\bibitem{steno-classification1}
Montalbo, F.J.P., Barfeh, D.P.Y.: Classification of stenography using
  convolutional neural networks and canny edge detection algorithm. In: 2019
  International Conference on Computational Intelligence and Knowledge Economy
  (ICCIKE). pp. 305--310 (2019)

\bibitem{steno-classification2}
Padilla, D.A., Vitug, N.K.U., Marquez, J.B.S.: Deep learning approach in gregg
  shorthand word to english-word conversion. In: 2020 IEEE 5th International
  Conference on Image, Vision and Computing (ICIVC). pp. 204--210 (2020)

\bibitem{pytorch}
Paszke, A., Gross, S., Massa, F., Lerer, A., Bradbury, J., Chanan, G., Killeen,
  T., Lin, Z., Gimelshein, N., Antiga, L., Desmaison, A., Kopf, A., Yang, E.,
  DeVito, Z., Raison, M., Tejani, A., Chilamkurthy, S., Steiner, B., Fang, L.,
  Bai, J., Chintala, S.: {PyTorch: An Imperative Style, High-Performance Deep
  Learning Library}. In: Wallach, H., Larochelle, H., Beygelzimer, A., d'Alché
  Buc, F., Fox, E., Garnett, R. (eds.) Advances in Neural Information
  Processing Systems 32. pp. 8024--8035. Curran Associates, Inc. (2019)

\bibitem{pylaia}
Puigcerver, J.: Are multidimensional recurrent layers really necessary for
  handwritten text recognition? In: 2017 14th IAPR International Conference on
  Document Analysis and Recognition (ICDAR). vol.~01, pp. 67--72 (2017)

\bibitem{htr-best-practices}
Retsinas, G., Sfikas, G., Gatos, B., Nikou, C.: Best practices for a
  handwritten text recognition system. In: Uchida, S., Barney, E., Eglin, V.
  (eds.) Document Analysis Systems. pp. 247--259. Springer International
  Publishing, Cham (2022)

\bibitem{wbs}
Scheidl, H., Fiel, S., Sablatnig, R.: Word beam search: A connectionist
  temporal classification decoding algorithm. In: 2018 16th International
  Conference on Frontiers in Handwriting Recognition (ICFHR). pp. 253--258
  (2018)

\bibitem{elastic}
Simard, P.Y., Steinkraus, D., Platt, J.C.: Best practices for convolutional
  neural networks applied to visual document analysis. In: Proceedings of the
  Seventh International Conference on Document Analysis and Recognition -
  Volume 2. p.~958. ICDAR '03, IEEE Computer Society, USA (2003)

\bibitem{flor}
de~{Sousa Neto}, A.F., {Leite Dantas Bezerra}, B., {Hector Toselli}, A.,
  {Baptista Lima}, E.: A robust handwritten recognition system for learning on
  different data restriction scenarios. Pattern Recognition Letters
  \textbf{159},  232--238 (2022)

\bibitem{torchvision}
{TorchVision maintainers and contributors}: {TorchVision: PyTorch's Computer
  Vision library} (11 2016)

\bibitem{khmer}
Valy, D., Verleysen, M., Chhun, S.: Data augmentation and text recognition on
  khmer historical manuscripts. In: 2020 17th International Conference on
  Frontiers in Handwriting Recognition (ICFHR). pp. 73--78 (2020)

\bibitem{rostock}
Wick, C., Z{\"o}llner, J., Gr{\"u}ning, T.: Rescoring sequence-to-sequence
  models for text line recognition with ctc-prefixes. In: Uchida, S., Barney,
  E., Eglin, V. (eds.) Document Analysis Systems. pp. 260--274. Springer
  International Publishing, Cham (2022)

\bibitem{distortion-aug}
Wigington, C., Stewart, S., Davis, B., Barrett, B., Price, B., Cohen, S.: Data
  augmentation for recognition of handwritten words and lines using a cnn-lstm
  network. In: 2017 14th IAPR International Conference on Document Analysis and
  Recognition (ICDAR). vol.~01, pp. 639--645 (2017)

\bibitem{wilcoxon}
Wilcoxon, F.: Individual comparisons by ranking methods. Biometrics Bulletin
  \textbf{1}(6),  80--83 (1945)

\bibitem{tomas}
Wilkinson, T., Brun, A.: Semantic and verbatim word spotting using deep neural
  networks. In: 2016 15th International Conference on Frontiers in Handwriting
  Recognition (ICFHR). pp. 307--312 (2016)

\bibitem{ctrlf}
Wilkinson, T., Lindstrom, J., Brun, A.: Neural ctrl-f: Segmentation-free
  query-by-string word spotting in handwritten manuscript collections. In:
  Proceedings of the IEEE International Conference on Computer Vision (ICCV)
  (Oct 2017)

\bibitem{gregg1916}
Zhai, F., Fan, Y., Verma, T., Sinha, R., Klakow, D.: A dataset and a novel
  neural approach for optical gregg shorthand recognition. In: International
  Conference on Text, Speech, and Dialogue. pp. 222--230. Springer (2018)

\end{thebibliography}


\begin{thebibliography}{1}
\providecommand{\url}[1]{\texttt{#1}}
\providecommand{\urlprefix}{URL }
\providecommand{\doi}[1]{https://doi.org/#1}

\bibitem{ctc_original}
Graves, A., Fern\'{a}ndez, S., Gomez, F., Schmidhuber, J.: Connectionist
  temporal classification: Labelling unsegmented sequence data with recurrent
  neural networks. In: Proceedings of the 23rd International Conference on
  Machine Learning. p. 369–376. ICML '06, ACM, New York, NY, USA (2006)

\bibitem{ctc_liwicki}
Graves, A., Liwicki, M., Fernández, S., Bertolami, R., Bunke, H., Schmidhuber,
  J.: A novel connectionist system for unconstrained handwriting recognition.
  IEEE Transactions on Pattern Analysis and Machine Intelligence
  \textbf{31}(5),  855--868 (2009)

\bibitem{adamw}
Loshchilov, I., Hutter, F.: Decoupled weight decay regularization. In:
  International Conference on Learning Representations (2019)

\end{thebibliography}
\end{document}

\title{Supplementary Material for:\\\enquote{A Study of Augmentation Methods for Handwritten Stenography Recognition}}
\titlerunning{Supplementary Material}

% AUTHORS AND INSTITUTIONS COMMENTED OUT FOR ANONYMITY
\author{Raphaela Heil\orcidID{0000-0002-5010-9149} \and
Eva Breznik\orcidID{0000-0003-3147-5626} }
%Third Author\inst{3}\orcidID{2222--3333-4444-5555}}
%%
\authorrunning{R. Heil, E. Breznik}
%% First names are abbreviated in the running head.
%% If there are more than two authors, 'et al.' is used.
%%
\institute{Centre for Image Analysis,\\ Department of Information Technology, \\Uppsala University, Uppsala, Sweden\\
\email{\{firstname\}.\{lastname\}@it.uu.se}\\
}
\maketitle              % typeset the header of the contribution

\section{Neural Network Training Protocol}
The model is trained with a batch size of eight, using a CTC-loss \cite{ctc_original,ctc_liwicki} and the AdamW \cite{adamw} optimiser with a learning rate of 0.001. After each training epoch, the validation loss is calculated and the model weights with the lowest validation loss are preserved for the final evaluation. Training is carried on for a maximum of 100 epochs, stopping earlier if the lowest validation loss has not been exceeded within the five most recent epochs.

\section{Visualisation of Evaluated Augmentations}
The following sections show examples of the extremes for the respective augmentation parameter range. For ease of comparison, all images are scaled to the same height. Padding and masking are visualised in grey, to emphasise the effect of the respective augmentation.

\subsection{Baseline}
\begin{figure}[h]
    \centering
    \includegraphics[height=16px]{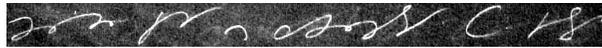}
    \caption{Baseline sample image (augmentation-free)}
    \label{fig:baseline-vis}
\end{figure}

\FloatBarrier
\clearpage

\subsection{Rotations}

\begin{figure}[h]
    \centering
    \begin{subfigure}[b]{0.45\textwidth}
         \centering
    \includegraphics[height=16px]{figures/aug_vis/flor_lower.png}
         \caption{-1.5 degrees}
     \end{subfigure}
     \hfill
     \begin{subfigure}[b]{0.45\textwidth}
         \centering
    \includegraphics[height=16px]{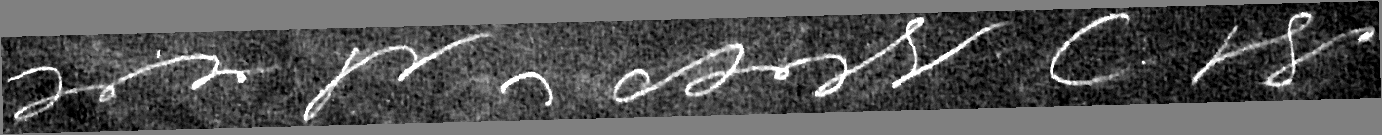}
         \caption{+1.5 degrees}
     \end{subfigure}
    \caption{rot1.5, negative and positive}
\end{figure}

\begin{figure}[h]
    \centering
    \begin{subfigure}[b]{0.45\textwidth}
         \centering
    \includegraphics[height=16px]{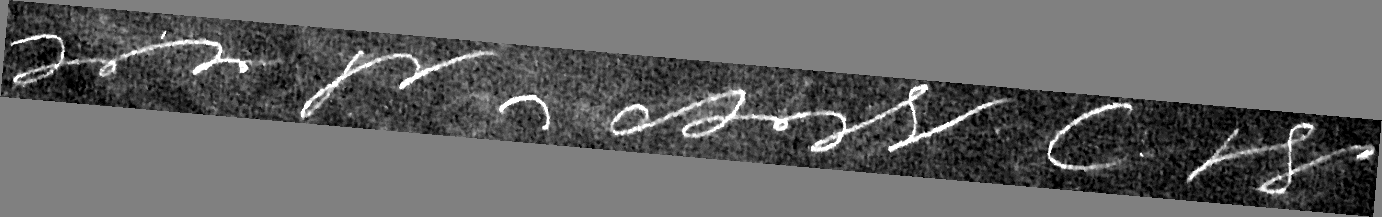}
         \caption{-5 degrees}
     \end{subfigure}
     \begin{subfigure}[b]{0.45\textwidth}
         \centering
    \includegraphics[height=16px]{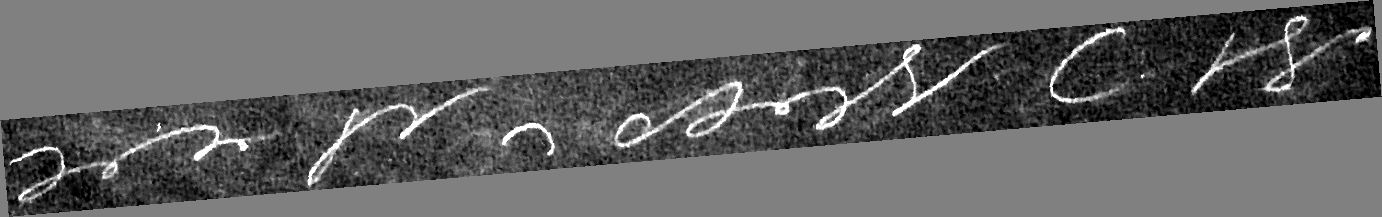}
         \caption{+5 degrees}
     \end{subfigure}
    \caption{rot5}
\end{figure}

\begin{figure}[h]
    \centering
    \includegraphics[height=16px]{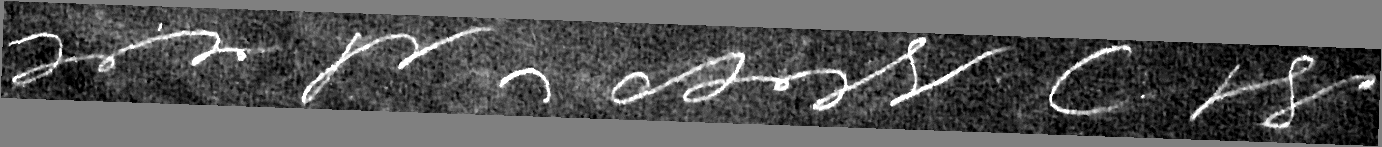}
    \caption{rot-2}
\end{figure}

\begin{figure}[h!]
    \centering
    \includegraphics[height=16px]{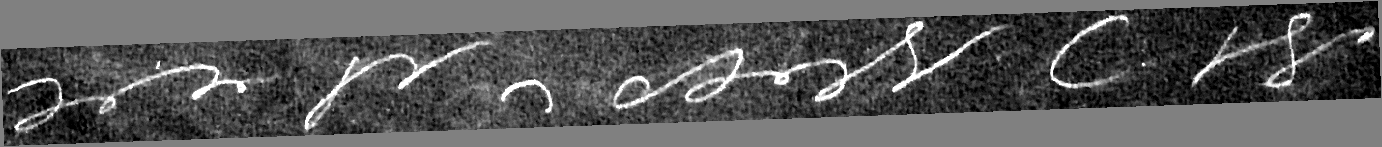}
    \caption{rot+2}
\end{figure}

\begin{figure}[h!]
    \centering
    \begin{subfigure}[b]{0.45\textwidth}
         \centering
    \includegraphics[height=16px]{figures/aug_vis/extreme_lower.png}
         \caption{-10 degrees}
     \end{subfigure}
     \begin{subfigure}[b]{0.45\textwidth}
         \centering
    \includegraphics[height=16px]{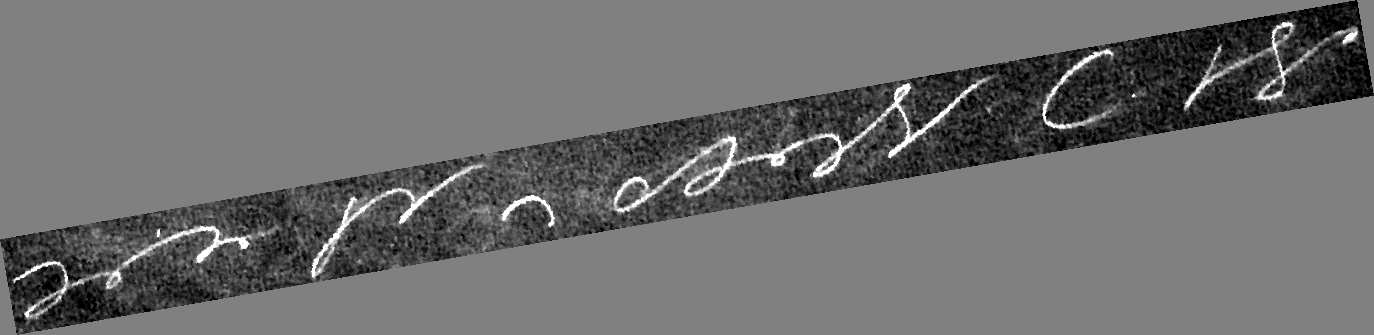}
         \caption{+10 degrees}
     \end{subfigure}
    \caption{rot10}
\end{figure}

\FloatBarrier
\clearpage

\subsection{Morphological Operations}
\begin{figure}[h!]
    \centering
    \begin{subfigure}[b]{0.9\textwidth}
         \centering
    \includegraphics[height=16px]{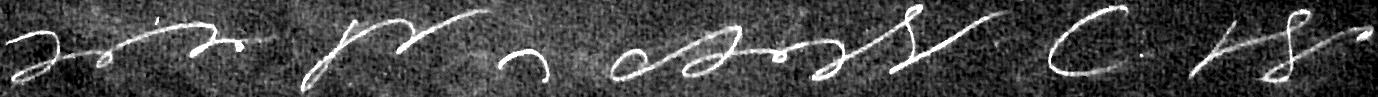}
         \caption{square structuring element, width=1}
     \end{subfigure}
     \begin{subfigure}[b]{0.9\textwidth}
         \centering
    \includegraphics[height=16px]{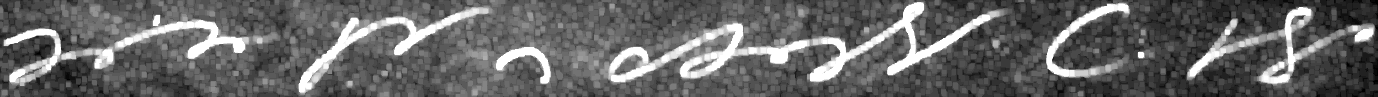}
         \caption{square structuring element, width=4}
     \end{subfigure}
     \begin{subfigure}[b]{0.9\textwidth}
         \centering
    \includegraphics[height=16px]{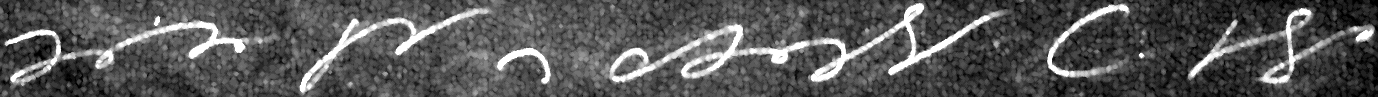}
         \caption{disk structuring element, radius=1}
     \end{subfigure}
     \begin{subfigure}[b]{0.9\textwidth}
         \centering
    \includegraphics[height=16px]{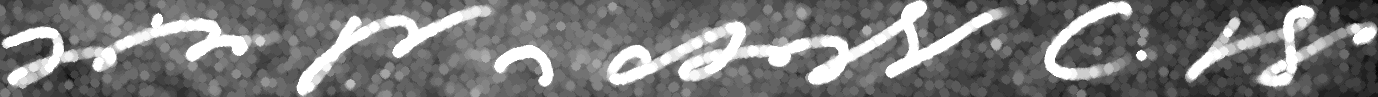}
         \caption{disk structuring element, radius=4}
     \end{subfigure}
    \caption{square-dilation and disk-dilation}
\end{figure}

\begin{figure}[h!]
    \centering
    \begin{subfigure}[b]{0.9\textwidth}
         \centering
    \includegraphics[height=16px]{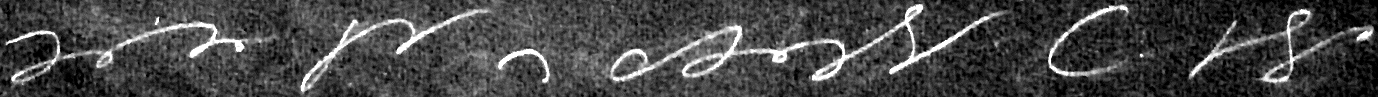}
         \caption{square structuring element, width=1}
     \end{subfigure}
     \begin{subfigure}[b]{0.9\textwidth}
         \centering
    \includegraphics[height=16px]{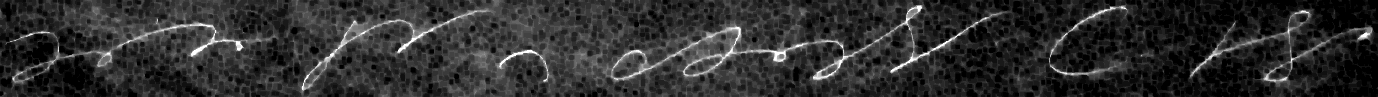}
         \caption{square structuring element, width=4}
     \end{subfigure}
     \begin{subfigure}[b]{0.9\textwidth}
         \centering
    \includegraphics[height=16px]{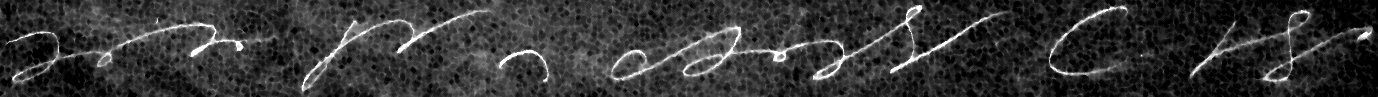}
         \caption{disk structuring element, radius=1}
     \end{subfigure}
     \begin{subfigure}[b]{0.9\textwidth}
         \centering
    \includegraphics[height=16px]{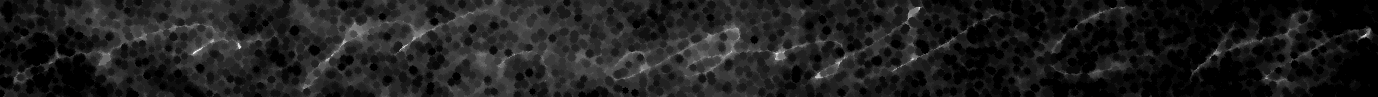}
         \caption{disk structuring element, radius=4}
     \end{subfigure}
    \caption{square-erosion and disk-erosion}
\end{figure}

\FloatBarrier
\clearpage
\subsection{Geometric Augmentations}

\begin{figure}[h!]
    \centering
    \begin{subfigure}[b]{0.9\textwidth}
         \centering
    \includegraphics[height=16px]{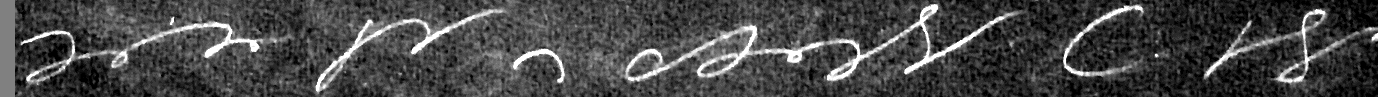}
         \caption{horizontal shift, 15px}
     \end{subfigure}
     \begin{subfigure}[b]{0.9\textwidth}
         \centering
    \includegraphics[height=16px]{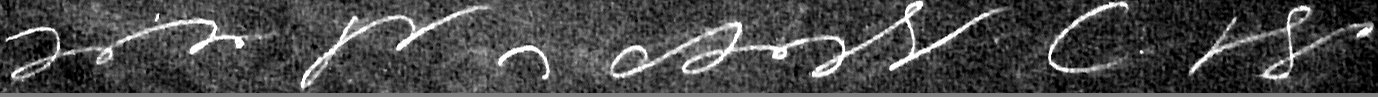}
         \caption{vertical shift, -3.5 px}
     \end{subfigure}
     \begin{subfigure}[b]{0.9\textwidth}
         \centering
    \includegraphics[height=16px]{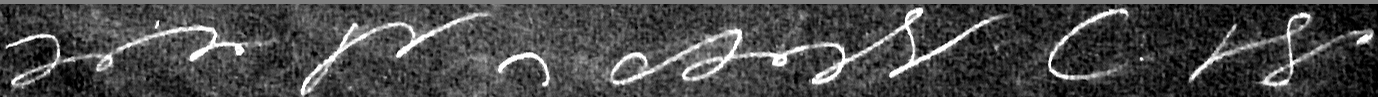}
         \caption{vertical shift, +3.5 px}
     \end{subfigure}
     \begin{subfigure}[b]{0.9\textwidth}
         \centering
    \includegraphics[height=16px]{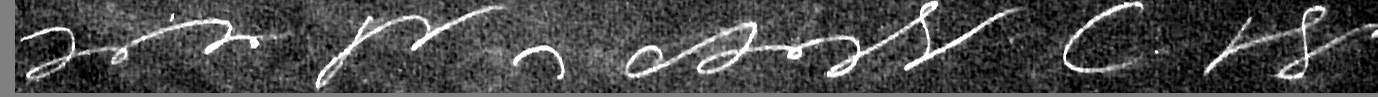}
         \caption{horizontal shift = 15px, vertical shift = -3.5 px}
     \end{subfigure}
          \begin{subfigure}[b]{0.9\textwidth}
         \centering
    \includegraphics[height=16px]{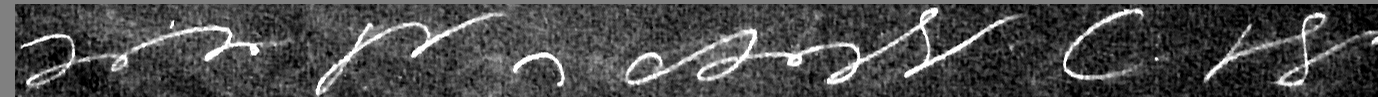}
         \caption{horizontal shift = 15px, vertical shift = +3.5 px}
     \end{subfigure}
    \caption{shift}
\end{figure}

\begin{figure}[h!]
    \centering
    \begin{subfigure}[b]{0.9\textwidth}
         \centering
    \includegraphics[height=16px]{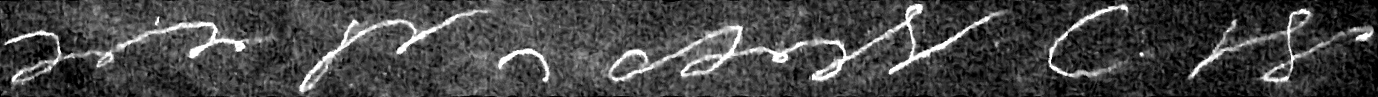}
         \caption{$\alpha=16$, $\sigma=5$}
     \end{subfigure}
     \begin{subfigure}[b]{0.9\textwidth}
         \centering
    \includegraphics[height=16px]{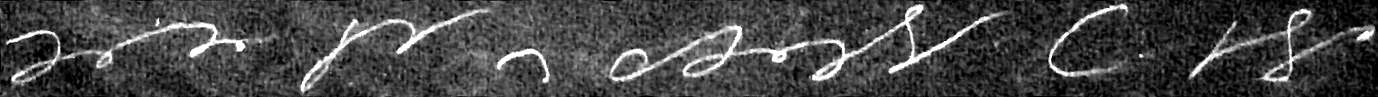}
                  \caption{$\alpha=16$, $\sigma=7$}
     \end{subfigure}
     \begin{subfigure}[b]{0.9\textwidth}
         \centering
    \includegraphics[height=16px]{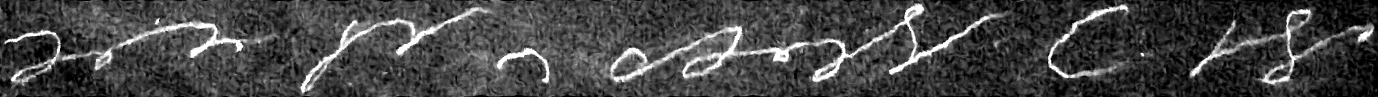}
                  \caption{$\alpha=20$, $\sigma=5$}
     \end{subfigure}
     \begin{subfigure}[b]{0.9\textwidth}
         \centering
    \includegraphics[height=16px]{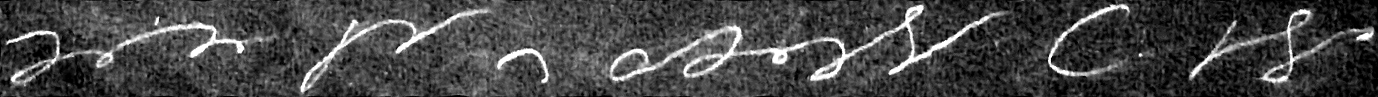}
                  \caption{$\alpha=20$, $\sigma=7$}
     \end{subfigure}
       
    \caption{elastic}
\end{figure}

\begin{figure}[h!]
    \centering
    \begin{subfigure}[b]{0.9\textwidth}
         \centering
    \includegraphics[height=16px]{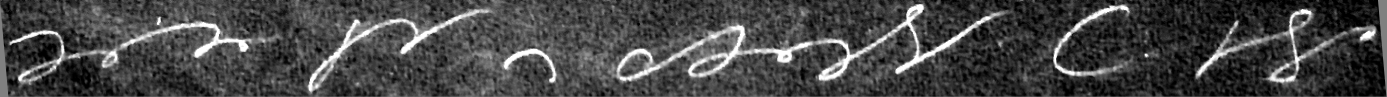}
         \caption{-5 degrees}
     \end{subfigure}
     \begin{subfigure}[b]{0.9\textwidth}
         \centering
    \includegraphics[height=16px]{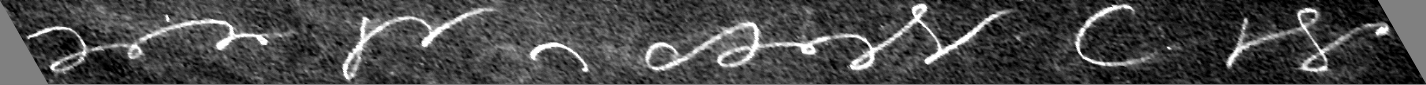}
                  \caption{-30 degrees}
     \end{subfigure}
     \begin{subfigure}[b]{0.9\textwidth}
         \centering
    \includegraphics[height=16px]{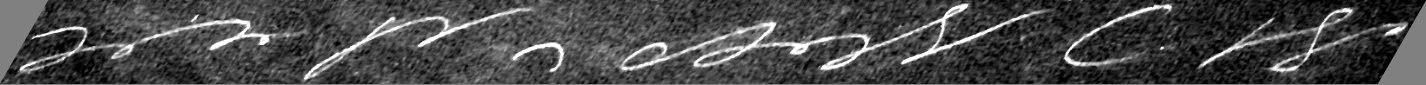}
                  \caption{+30 degrees}
     \end{subfigure}
    \caption{shear and shear30}
\end{figure}

\begin{figure}[h!]
    \centering
    \begin{subfigure}[b]{0.9\textwidth}
         \centering
    \includegraphics[height=16px]{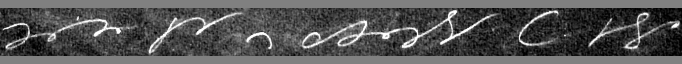}
         \caption{scale factor = 0.75}
     \end{subfigure}
     \begin{subfigure}[b]{0.9\textwidth}
         \centering
    \includegraphics[height=16px]{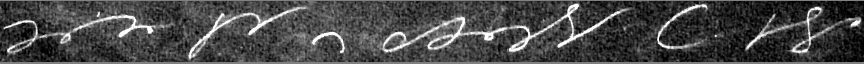}
                  \caption{scale factor = 0.95}
     \end{subfigure}
    \caption{scale75 and scale95}
\end{figure}

\FloatBarrier
%\clearpage
\subsection{Intensity Augmentations}

\begin{figure}[!htbp]
    \centering
    \begin{subfigure}[b]{0.9\textwidth}
         \centering
    \includegraphics[height=16px]{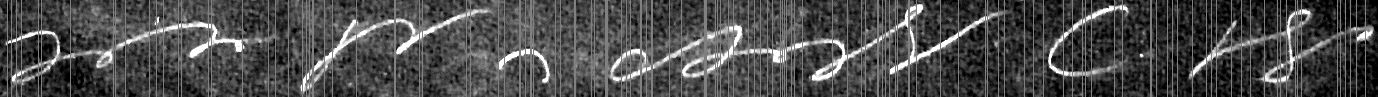}
         \caption{10\% of columns masked}
     \end{subfigure}
     \begin{subfigure}[b]{0.9\textwidth}
         \centering
    \includegraphics[height=16px]{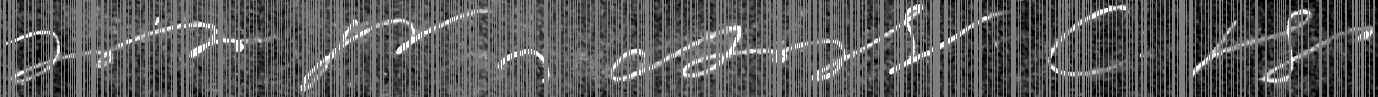}
                  \caption{40\% of columns masked}
     \end{subfigure}
    \caption{mask and mask40}
\end{figure}

\begin{figure}[!htbp]
    \centering
    \begin{subfigure}[b]{0.9\textwidth}
         \centering
    \includegraphics[height=16px]{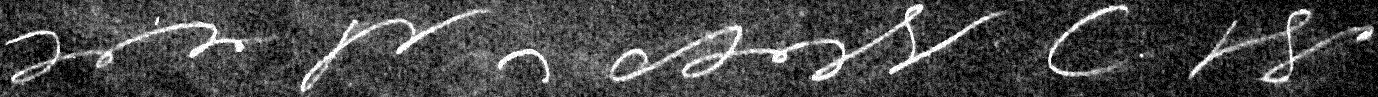}
         \caption{$\sigma=0.08$}
     \end{subfigure}
     \begin{subfigure}[b]{0.9\textwidth}
         \centering
    \includegraphics[height=16px]{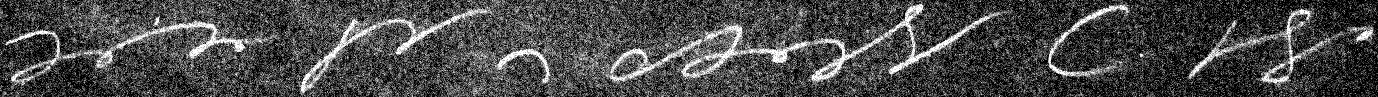}
                  \caption{$\sigma=0.18$}
     \end{subfigure}
    \caption{noise}
\end{figure}

\begin{figure}[h]
    \centering
    \includegraphics[height=16px]{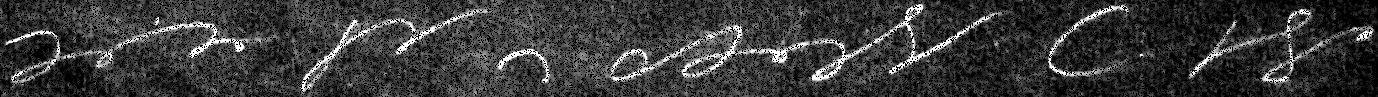}
    \caption{dropout, 20\% of pixels masked}
\end{figure}

\begin{figure}[h]
    \centering
    \begin{subfigure}[b]{0.9\textwidth}
         \centering
    \includegraphics[height=16px]{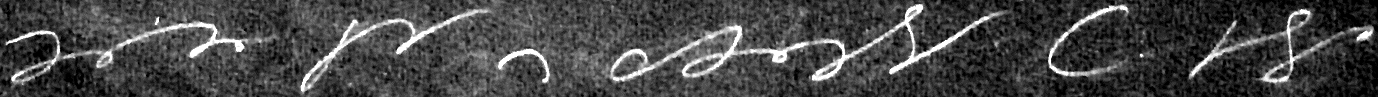}
         \caption{$\sigma=0.1$}
     \end{subfigure}
     \begin{subfigure}[b]{0.9\textwidth}
         \centering
    \includegraphics[height=16px]{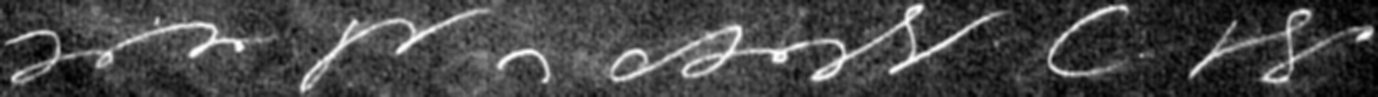}
                  \caption{$\sigma=2.0$}
     \end{subfigure}
    \caption{blur}
\end{figure}

\bibliographystyle{splncs04}
\bibliography{main}